\documentclass[letterpaper]{article} 
\usepackage{aaai2026}  
\usepackage{times}  
\usepackage{helvet}  
\usepackage{courier}  
\usepackage[hyphens]{url}  
\usepackage{graphicx} 
\usepackage{natbib}  
\usepackage{caption} 
\usepackage{algorithm}
\usepackage{algorithmic}

\usepackage{newfloat}
\usepackage{listings}

\usepackage{bibentry}

\usepackage{bibentry}
\usepackage{amsmath}
\usepackage{amsfonts}
\usepackage{booktabs}
\usepackage{multirow}
\usepackage{array}

\DeclareCaptionStyle{ruled}{labelfont=normalfont,labelsep=colon,strut=off} 
\floatstyle{ruled}
\newfloat{listing}{tb}{lst}{}
\floatname{listing}{Listing}
\title{Towards High-Fidelity, Identity-Preserving Real-Time Makeup Transfer: Decoupling Style Generation}
\author{
    Lydia Kin Ching Chau\equalcontrib,   Zhi Yu\equalcontrib,  Ruowei Jiang
}
\affiliations{
    kinchinglydia.chau@loreal.com,  vicky.yu@loreal.com,    irene.jiang2@loreal.com\\
    Modiface Inc.
}

\begin{document}
\maketitle
\begin{abstract}
We present a novel framework for real-time virtual makeup try-on that achieves high-fidelity, identity-preserving cosmetic transfer with robust temporal consistency. In live makeup transfer applications, it is critical to synthesize temporally coherent results that accurately replicate fine-grained makeup and preserve user's identity. However, existing methods often struggle to disentangle semitransparent cosmetics from skin tones and other identify features, causing identity shifts and raising fairness concerns. Furthermore, current methods lack real-time capabilities and fail to maintain temporal consistency, limiting practical adoption. To address these challenges, we decouple makeup transfer into two steps: transparent makeup mask extraction and graphics-based mask rendering. After the makeup extraction step, the makeup rendering can be performed in real time, enabling live makeup try-on. Our makeup extraction model trained on pseudo-ground-truth data generated via two complementary methods: a graphics-based rendering pipeline and an unsupervised k-means clustering approach. To further enhance transparency estimation and color fidelity, we propose specialized training objectives, including alpha-weighted reconstruction and lip color losses. Our method achieves robust makeup transfer across diverse poses, expressions, and skin tones while preserving temporal smoothness. Extensive experiments demonstrate that our approach outperforms existing baselines in capturing fine details, maintaining temporal stability, and preserving identity integrity.
\end{abstract}


\section{Introduction}

\begin{figure*}[t]
    \begin{center}
        \includegraphics[width=1\textwidth]{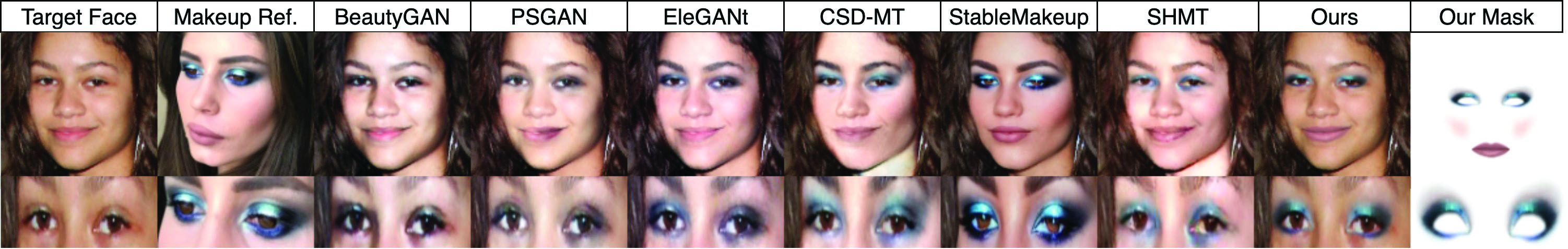}
        \caption*{(a) Comparison of makeup transfer across skin tones using our method and prior methods\cite{beautygan, psgan, elegant, csd-mt, stableMakeup, sun2024shmt}.}
        \includegraphics[width=1\textwidth]{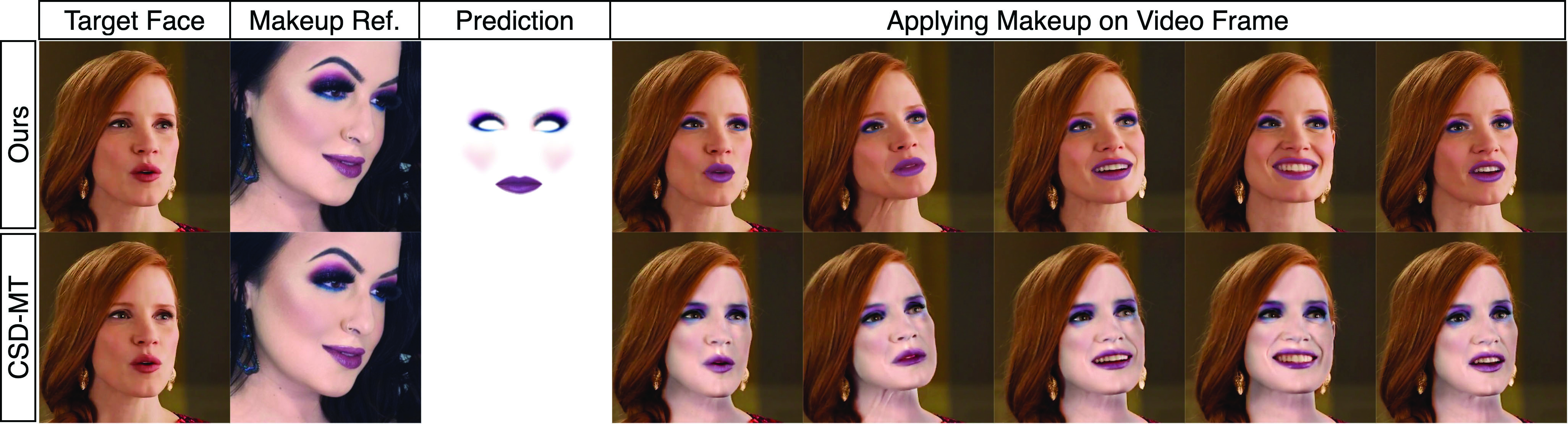}
        \caption*{(b) Makeup transfer on video frames.}
    \caption{Our method extracts high-fidelity RGBA masks that disentangle makeup from skin tone and identity features, enabling transfer across skin tones. Our inference pipeline decouples makeup mask extraction and application, enabling temporally consistent makeup virtual try-on in videos.}
    \label{fig:front_page_vids}
    \end{center}
\end{figure*}
Virtual makeup try-on (VTO) systems enable users to visualize cosmetic styles on their faces through images, videos, or live demonstrations. Example-based makeup transfer\cite{example-basedCosmeticTransfer} has become a dominant approach, aiming to synthesize photorealistic results that replicate the fine-grained appearance of a reference makeup while preserving the subject’s identity. In practical applications, temporal consistency and stability are also essential, especially or video-based or real-time settings.

Early methods used graphics-based rendering \cite{example-basedCosmeticTransfer, physics-based, digitalFaceByExample, xu2013automatic}, which can produce high-quality results under controlled conditions. However, they depend on strict assumptions about lighting, pose, and expression alignment between before-and-after makeup pairs, limiting their effectiveness in real-world, unconstrained settings. Moreover, these methods are not suitable for live makeup transfer.

Recent advances in conditional GANs \cite{pix2pix} have enabled deep learning-based makeup transfer. Due to the lack of large paired datasets, weakly supervised methods have gained popularity. \citet{pairedcyclegan} proposed a dual-model framework with cycle-consistency losses \cite{cyclegan} to learn from unpaired data but struggle to capture fine-grained details. To improve supervision, studies have explored pseudo-ground-truth generation \cite{ladn, SCGAN, ssat, zhang2019disentangled, zhu2022semi, beautyrec}, including geometric warping \cite{facialTransformersTps, elegant}, histogram matching \cite{beautygan, elegant, psgan}, and synthetic overlays \cite{lipstickArentEnough}. However, these methods often assume consistent color distributions, overlooking the semi-transparent nature of makeup, which varies with coating thickness, absorption, and reflection \cite{kubelka1931beitrag}, making disentanglement from skin tone challenging.

Additionally, due to the high computational cost of high-fidelity makeup transfer, existing methods do not support live applications. The closest approach, proposed by \cite{tinyBeauty}, allows real-time makeup application but is limited to a small manually predefined set of styles, limiting the flexibility and broader applicability of makeup transfer.

To disentangle semi-transparent makeup from underlying surfaces, we propose a novel data generation pipeline that produces makeup images paired with ground-truth semi-transparency masks capturing only makeup content without identity-related features. Our approach combines two complementary methods: a graphics rendering-based pipeline that enables controlled variation in makeup attributes, such as shape, color and transparency; and a k-means clustering technique to extract pseudo ground truth semi-transparent makeup masks from real images.

Building upon this data foundation, we introduce new training objectives and an inference pipeline to extract and apply transparency-sensitive makeup masks. Specifically, we design two loss terms to improve both the accuracy and smoothness of mask predictions. Our method enables the precise transfer of the makeup layer across various skin tones while preserving detailed color and shape characteristics. These identity-free, semi-transparent masks decouple the makeup extraction and application processes, allowing for real-time, temporally consistent virtual makeup try-on.
In summary, our contributions are as follows:
\begin{itemize}
\item We present a decoupled makeup transfer pipeline that separates makeup transparent mask extraction and application, enabling real-time and temporally consistent video inference.

\item We present a novel data generation pipeline for pseudo-ground-truth transparent makeup masks, enabling robust learning of makeup extraction and skin tone disentanglement. Our approach synergistically combines a graphics-based rendering engine with a diverse library of cosmetic styles and a K-means clustering method that extracts transparent masks from geometrically aligned real images.

\item We introduce new training objectives for makeup extraction that incorporates novel lip color and alpha-weighted reconstruction losses. Our model achieves unprecedented color and shape fidelity while robustly disentangling makeup from underlying skin tones.
\end{itemize}

\begin{figure}[t!]
\begin{center}

\includegraphics[width=\columnwidth]{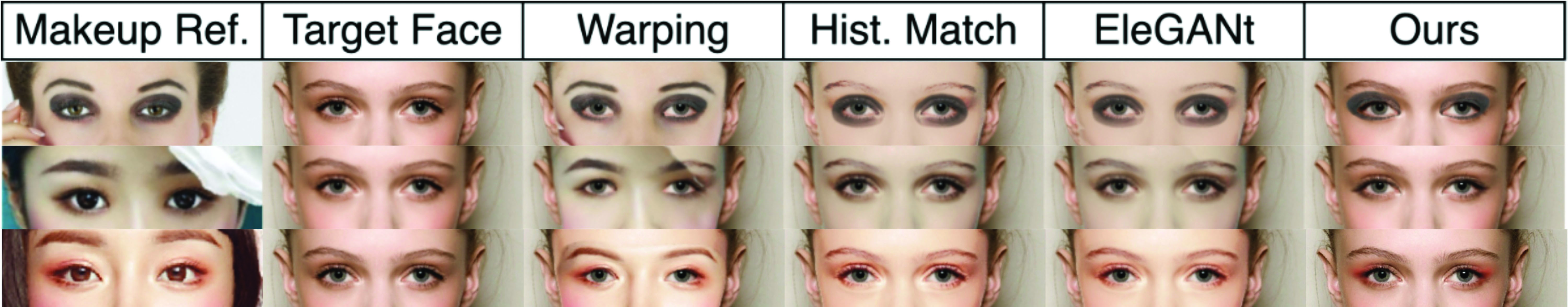}
\caption{Comparison of pseudo ground truth generation methods for eye makeup.}
\label{fig:dataComparison}
\end{center}
\end{figure}

\begin{figure}[t!]
\begin{center}
\includegraphics[width=\columnwidth]{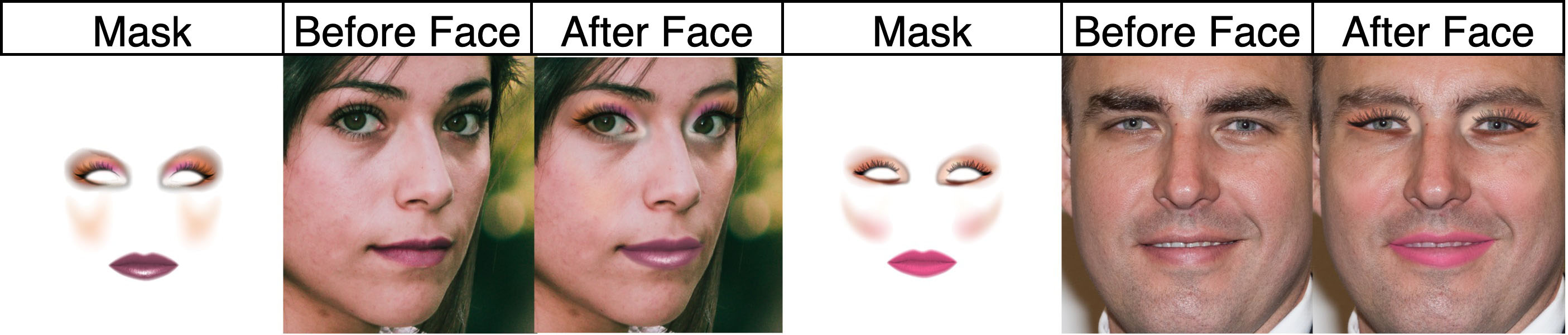}
\caption{Graphic-rendering pseudo ground truth pipeline that combines makeup masks with natural faces to create after makeup image. The face images are from FFHQ dataset\cite{styleGAN_ffhq}.}
\label{fig:graphic_data}
\end{center}
\end{figure}

\begin{figure}[t!]
\begin{center}
\includegraphics[width=\columnwidth]{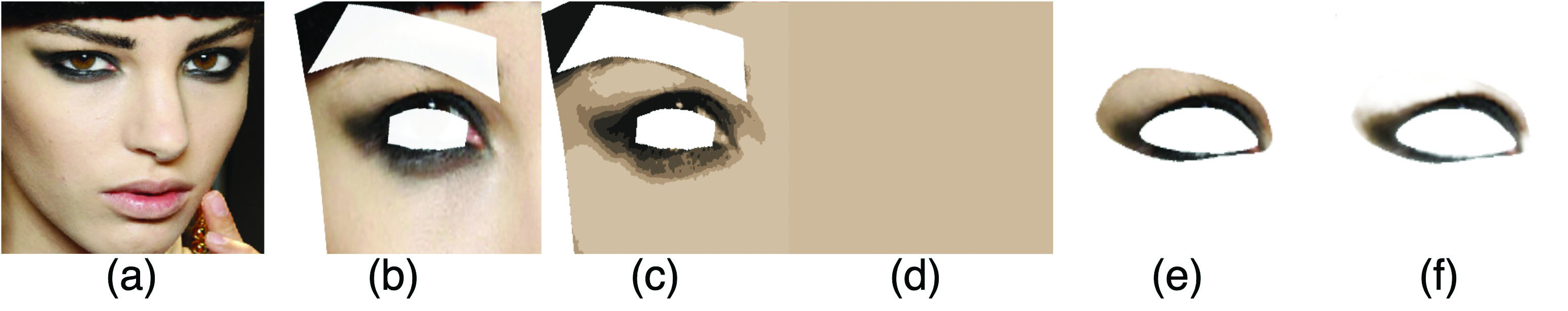}
\caption{Pipeline for generating an eye makeup mask using warping and the $k$-means clustering method. (a) Input makeup image. (b) Image warped to the canonical face layout, with inner eye and eyebrow regions removed. (c) Color clustering using $k$-means. (d) Skin tone estimation via the weighted average of the top-$s$ most frequent color clusters. (e) Application of the eye parsing mask to isolate the eyeshadow region. (f) RGBA mask generation, where the alpha channel is computed based on similarity to the estimated skin tone using Equation \ref{eq:kmean}.}
\label{fig:kmean_ill}
\end{center}
\end{figure}

\section{Related Work}
\subsection{Makeup Transfer Pseudo Ground Truth Data\label{subsection:pusedogt}}
Recent work in makeup transfer has focused on improving color fidelity and spatial alignment. A common approach involves pseudo-ground truth synthesis to supervise generative models, as shown in Figure \ref{fig:dataComparison}. \citet{beautygan} use histogram matching to enforce regional color transfer, which works well for simple cases like lipstick but struggles with complex spatial patterns such as eye makeup. \citet{facialTransformersTps} propose warping makeup references to align with target faces, enabling better texture transfer but requiring precise alignment, which is prone to artifacts under facial variation. \citet{elegant} combine warping with histogram matching to balance their strengths. \citet{lipstickArentEnough} introduce a dataset of facial stickers to mimic special makeup effects. Despite these efforts, many methods still assume consistent color distributions between the makeup reference and target face, overlooking critical factors like transparency and skin tone variation.

Recently, large vision-language models (VLMs) have been explored for synthetic data generation. \citet{tinyBeauty} propose a pipeline to expand a limited set of makeup pairs into a larger before-and-after dataset. \citet{stableMakeup} use VLMs to generate makeup images from non-makeup faces guided by text prompts. Although these methods produce realistic results, they lack fine-grained control over key attributes such as color, transparency, finish, and shape across facial regions. This limitation stems from the inherent difficulty VLMs face in generating highly specific visual concepts\cite{anImgOneWord, Dreambooth}, restricting the diversity and fidelity of the resulting makeup styles.

In contrast, our proposed graphic rendering pipeline supports a broad range of makeup styles with controllable combinations of color, finish, and transparency. Also, built on a k-means-based approach, it requires only input makeup images to generate various makeup masks efficiently.
 
\subsection{Identity Preservation in Makeup Transfer}
Another critical aspect of makeup transfer is identity preservation, ensuring the transferred face is not influenced by the identity of the makeup reference. Many methods use identity loss\cite{beautygan}, typically implemented as perceptual loss\cite{perceptualLoss}, to enforce this. \citet{csd-mt} observe that makeup mainly affects low-frequency facial components and propose preserving high-frequency details to better maintain identity. While helpful at preserving some facial structure, these approaches often neglect skin tone consistency, a critical aspect of identity.

To control skin tone, some works apply regional shade editing. \citet{elegant} segment the face into skin, lips, and eyes for localized feature interpolation while \citet{SCGAN} decompose makeup into regional style codes. Though effective, these methods depend on external facial parsing and struggle with semi-transparent makeup where skin tone subtly affects makeup appearance.

To overcome this, we propose introducing transparent makeup mask extraction as an intermediate step in the transfer pipeline. Our model explicitly regresses alpha values, enabling disentanglement between the makeup layer and the skin tone, thus preserving identity and appearance realism.

\section{Method}

\begin{figure*}[t!]
\begin{center}
\includegraphics[width=0.9\textwidth]{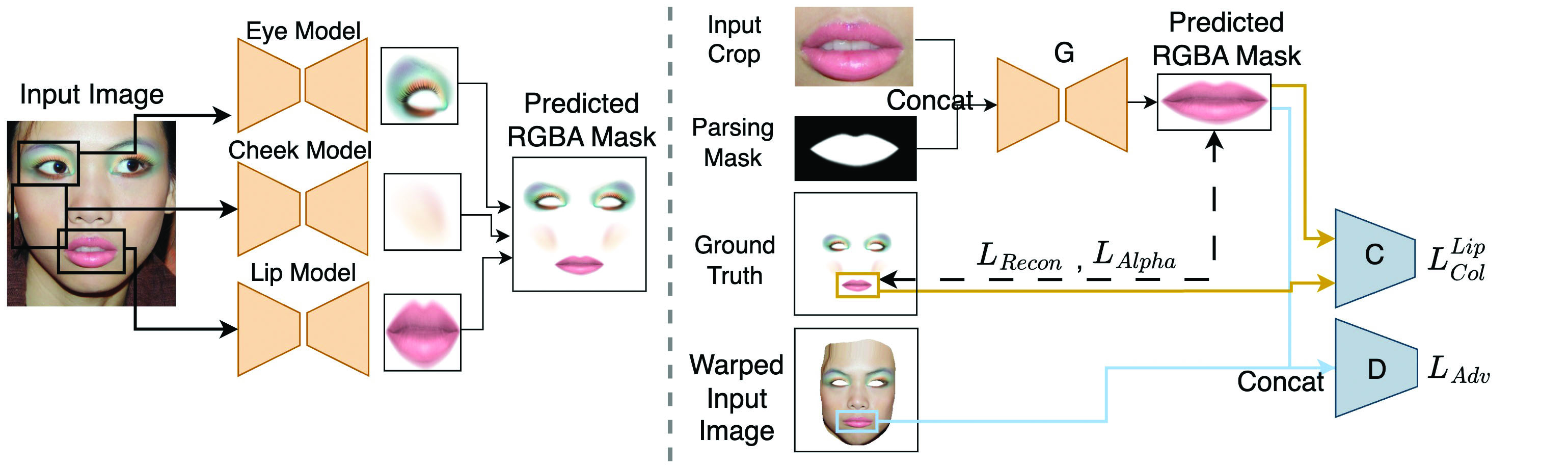}
\caption{Left: Model architecture during inference. Right: Training architecture of the lip model.} 
\label{fig:modelArchitecture}
\end{center}
\end{figure*}
\subsection{Graphics-Rendered Pseudo Ground Truth} \label{subsection:grphicdata}
We generate paired data consisting of transparent makeup masks and corresponding makeup-applied images by first synthesizing diverse masks from a comprehensive library of makeup styles, and then warping these masks onto non-makeup face images. To construct this library, we collect a wide range of shape templates for various makeup components, including eyeshadow, eyeliner, blush and lipstick. As illustrated in Figure \ref{fig:graphic_data}, we then randomly sample shapes for individual makeup regions and vary their transparency, color and finish. These components are then composited onto a canonical face layout to form full-face makeup masks. This procedural generation enables effectively infinite combinations of makeup styles, while ensuring the masks remain disentangled from facial identity.

To create the final paired data, we sample a synthesized mask for each natural face image and apply it using 2D thin-plate spline warping\cite{tps_warp}, followed by alpha blending. This process produces a realistic makeup-applied image paired with its corresponding transparent mask. The resulting dataset serves as pseudo ground truth for supervised training, enabling scalable and diverse supervision without requiring real before-and-after makeup pairs.

\subsection{Warping and K-Means for Eye Pseudo Labels\label{subsection:kmean}}
While the generation of graphics-based makeup data enables the creation of high-quality supervised training data that perfectly disentangle makeup and identity, it is inherently limited by the set of predefined shape templates. This limitation is particularly significant for eye makeup, where there is often considerable variation in the shapes of eyeshadow and eyeliner. To further increase the diversity and realism of makeup styles, we introduce an unsupervised method for generating partial eye makeup masks directly from real makeup images.

As illustrated in Figure \ref{fig:kmean_ill}, we first warp the makeup image $x$ to a canonical face layout. We then estimate the base skin tone by computing a weighted average of the top-$s$ most frequent colors, using a $k$-means clustering algorithm over the pixel color distribution. Based on the assumption that makeup becomes less transparent as its color deviates further from the skin tone, we estimate per-pixel alpha values using the inverse of cosine similarity between the pixel color and the skin tone.
{\small
\begin{equation}
\text{Alpha}_{i,j} = \max\left(0,\ 1 - \text{CosSim}(x^{\text{LAB}}_{i,j},\ \text{skinTone}^{\text{LAB}})\right)
\label{eq:kmean}
\end{equation}
}
where $x^{LAB}_{i,j}$ is the pixel at location (i,j) and $skinTone^{LAB}$ is the estimated skin tone, both in LAB color space. We set $k=6, s=2$ , to capture the range of potential makup color clusters near eye regions. These estimated alpha values are utilized as weights to guide the learning of the transparent mask, as outlined in Section \ref{subsec:lossfunction}.

\subsection{Model Architecture}
As illustrated in Figure \ref{fig:modelArchitecture}, the framework comprises individual models to extract eye, lip, and cheek makeup; each extraction model is paired with a corresponding discriminator, and the lip model includes an additional color regressor. We adapt the UNet-based generator and discriminator architecture from \citet{pix2pix}, incorporating several modifications to suit the makeup extraction task.

To align the generated makeup masks with the location of the canonical face, the generator ($G$) is conditioned on the concatenation of the makeup reference image ($x$) and an average alpha mask. The average alpha mask is created by averaging the alpha channel of the training data. To mitigate mode collapse\cite{conditionalGAN}, the discriminator ($D$) is conditioned on the reference makeup image, warped to the canonical face location. To improve the accuracy of lip makeup color, we introduce a lip color regressor ($C$) that outputs the average RGB values of a given lip makeup mask. See supplementary material for more details.

\subsection{Full Objective\label{subsec:lossfunction}}
We train the generators and discriminators jointly, minimizing a loss function that includes reconstruction loss ($L^{p}_{Recon}$), adversarial losses ($L^{p}_{Adv_G}$, $L^{p}_{Adv_D}$), and alpha loss ($L^{p}_{Alpha}$). For the lip model, we add a color regression loss ($L^{Lip}_{Col}$) to improve color matching. The total loss is:

{\small
\begin{equation}
\begin{split}
L_{\text{Total}} = \sum_{p \in P} \lambda^{p}_{\text{Recon}} L^{p}_{\text{Recon}} 
+ \lambda^{p}_{\text{Alpha}} L^{p}_{\text{Alpha}} \\
+ \lambda^{\text{Lip}}_{\text{Col}} L^{\text{Lip}}_{\text{Col}}
+ \lambda^{p}_{\text{Adv}} (L^{p}_{\text{Adv\_G}} + L^{p}_{\text{Adv\_D}}) \\
\end{split}
\end{equation}
}
where $P=\{Lip, Eye, Cheek\}$ denotes the facial parts, and $\lambda$ are hyperparameter controlling the relative weights.

\textbf{Alpha-weighted reconstruction loss ($L_{Recon}$).}
The reconstruction loss is an L1 pixel-wise loss between the predicted mask $\hat{y}\in\{\mathbb{R}^{W*H*4}\}$ and ground truth mask $y\in\{\mathbb{R}^{W*H*4}\}$. To account for the transparency of the output mask, we weight the pixel-wise differences by the alpha channel ($\alpha$) of the ground truth makeup mask. 
{\small
\begin{equation}
L_{Recon}= \sum^W_{i=1}\sum^H_{j=1}\sum_{k\in\{R,G,B\}}y_{\alpha,i,j} * |\hat{y}_{k,i,j}-y_{k,i,j}|
\end{equation}
}
where $y_{k,i,j}$ and $\hat{y}_{k,i,j}$ denote the values of channel $k$ at pixel location $(i,j)$ in the ground truth mask and the predicted mask, respectively.

\textbf{Alpha loss ($L_{Alpha}$).} We apply a standard L1 pixel-wise loss between the alpha channels of the predicted and ground-truth masks, using ground-truth data generated from our graphics-based rendering pipeline.

\textbf{Lip color loss ($L_{Col}$).}
To improve lip color accuracy, prior to training the generator, we train a lip color regressor using graphic makeup mask data. Given a makeup mask, the regressor predicts the average RGB values of the lip pixels, $C:\mathbb{R}^{W*H*4}\rightarrow \mathbb{R}^3$. It is trained with the following objective.
{\small
\begin{equation}
L_{ColorRegressor}= \Big\|C(y \odot M)- \frac{1}{n}\sum^W_{i=1}\sum^H_{j=1}(y \odot M)\Big\|_2
\end{equation}
}
where $M\in\{0, 1\}^{W*H*4}$ is a binary lip segmentation mask repeated along the channel dimension, $n$ is the count of nonzero values in the $W*H$ binary lip mask, and $\odot$ denotes pixel-wise multiplication.

During training of the makeup extraction model, the regressor is frozen, and the color loss is computed as the mean squared difference between the average color of the predicted and ground truth masks:
{\small
\begin{equation}
L_{Col}^{Lip}= \Big\|C(\hat{y} \odot M)-C(y \odot M)\Big\|_2
\label{eq:L_col}
\end{equation}
}

\textbf{Adversarial losses ($L_{Adv_G}$, $L_{Adv_D}$).} We use standard cross-entropy adversarial loss. To prevent mode collapse, the discriminator is conditioned on the warped makeup reference. We warped the reference to the canonical face location to ensure better geometric alignment between the makeup mask and the reference image.

\subsection{Inference Pipeline}
The performance of UNet-based generative models relies on precise spatial alignment between input and output. However, makeup reference faces often vary in pose or shape from the canonical face. To address this, prior methods \cite{psgan, SCGAN, elegant, csd-mt, sun2024shmt} use external face parsing models during preprocessing to extract semantic regions like eyes, lips, and skin. These region maps then guide geometric normalization through facial region warping or generator conditioning.

We adopt a simpler approach by using an off-the-shelf lightweight face tracking model \cite{chau2024occlusion} to extract facial keypoints. An affine transformation is then applied to align the input based on the average positions of the eyes and lips. Without explicitly passing landmark information to the model, this lightweight preprocessing step standardizes facial geometry and enhances spatial correspondence between input and output, leading to higher-fidelity generation.

\textbf{Video Inference Pipeline.} We adopt a two-stage pipeline. First, our makeup extraction model generates a transparent makeup mask from a reference image. Then, for each video frame, facial landmarks and a face parsing mask are detected. The extracted makeup is warped onto the target face to simulate the desired look, while the parsing mask ensures that makeup is applied only within the facial region. Leveraging accurate, real-time facial alignment and segmentation models \cite{zhang2014coarse, ning2020real, mobilefan, chau2024occlusion}, along with a lightweight warping process, our pipeline enables real-time inference once the mask is extracted. As the same mask is consistently applied across frames, temporal consistency of the makeup style is inherently maintained throughout the video.

\subsection{Implementation Details}
We train the generator and discriminator networks jointly for 55 epochs with a batch size of 8, updating both at the end of each epoch. Training is conducted on an NVIDIA A100 GPU and takes approximately 12 hours. Prior to adversarial training, the color regressor is pretrained for 10 epochs and kept frozen thereafter, with batch size of 32 and the training takes around an hour. The generator and discriminator are optimized using the Adam optimizer\cite{adam} with $\beta_1 = 0.5$, $\beta_2 = 0.999$, and a learning rate of $2 \times 10^{-4}$, using two time-scale update rule\cite{Seitzer2020FID}. The color regressor is trained using stochastic gradient descent (SGD) with a learning rate of $5 \times 10^{-5}$ and momentum of 0.9. The loss weights are set as: $\lambda_{\text{Recon}} = 100$, $\lambda_{\text{Adv}} = 10$, $\lambda_{\text{Alpha}} = 100$, and $\lambda^{\text{Lip}}_{\text{Col}} = 50$.

During preprocessing, the eye, lip, and cheek regions are cropped based on the minimum and maximum coordinates of their facial landmarks. These regions are resized to $256 \times 256$, approximating a $1024 \times 1024$ full-face resolution. We apply standard data augmentation techniques, including random translation, scaling, and Gaussian blurring.

\begin{table}[t]
\begin{center}
\small
\setlength{\tabcolsep}{1mm}
\begin{tabular}{
    >{\raggedright\arraybackslash}p{1.49cm}
    >{\centering\arraybackslash}p{0.45cm}
    >{\centering\arraybackslash}p{0.68cm}
    >{\centering\arraybackslash}p{0.65cm}
    >{\centering\arraybackslash}p{0.8cm}
    >{\centering\arraybackslash}p{0.45cm}
    >{\centering\arraybackslash}p{0.68cm}
    >{\centering\arraybackslash}p{0.65cm}
    >{\centering\arraybackslash}p{0.7cm}
}
\toprule
\multirow{2}{*}{Model} & \multicolumn{4}{c}{LADN (\citeauthor{psgan})} & \multicolumn{4}{c}{Wild (\citeauthor{ladn})} \\
 & FID & LPIPS & PSNR & FID(I) & FID & LPIPS & PSNR & FID(I) \\
\midrule
\citeauthor{beautygan}    & 34.5 & 0.11 & 29.3 & 28.6 & 34.2 & 0.13 & 28.7 & 26.2 \\
\citeauthor{psgan}        & 29.3 & 0.16 & 28.8 & 20.7 & 34.4 & 0.14 & 28.6 & 23.2 \\
\citeauthor{elegant}      & 26.8 & 0.11 & 29.6 & 23.3 & 35.5 & 0.13 & 29.0 & 25.9 \\
\citeauthor{csd-mt}       & 29.1 & 0.12 & 31.3 & 35.1 & 28.1 & 0.14 & 31.0 & 27.4 \\
\citeauthor{stableMakeup} & 33.0 & 0.13 & 30.0 & 52.1 & 63.0 & 0.15 & 29.7 & 39.1 \\
\citeauthor{sun2024shmt}  & 37.7 & 0.20 & 30.0 & 35.0 & 42.4 & 0.23 & 29.3 & 35.6 \\

\textbf{Ours}         & \textbf{9.7} & \textbf{0.02} & \textbf{40.6} & \textbf{11.6} & \textbf{11.2} & \textbf{0.02} & \textbf{41.0} & \textbf{8.0} \\
\bottomrule
\end{tabular}
\caption{Quantitative evaluation of makeup transfer fidelity (FID$\downarrow$, LPIPS$\downarrow$, PSNR$\uparrow$) and identity preservation (FID(I)$\downarrow$). $\downarrow$ lower is better, $\uparrow$ higher is better.}
\label{tbl:FID}
\end{center}
\end{table}

\section{Experiment}
\subsection{Experimental Setting}
We train our model using natural face images from the FFHQ dataset\cite{styleGAN_ffhq} and makeup images from the Makeup Transfer (MT) dataset\cite{beautygan}. We randomly sample 20,000 makeup-free, occlusion-free images from FFHQ, generating three synthetic makeup masks per image based on the graphic rendering-based method. The MT dataset consists of 1,115 before-makeup and 2,719 after-makeup images, only the makeup images are used for training. These images cover a broad range of makeup styles from subtle to heavy. Pseudo ground-truth labels are generated using the warping and K-mean method.

For evaluation, we use three datasets: Makeup-Wild\cite{psgan}, LADN\cite{ladn}, and the non-makeup images from the MT dataset. Makeup-Wild dataset includes 384 before and 403 after-makeup images with diverse makeup styles, expressions, and head poses. LADN dataset comprises 334 before and 384 after-makeup images, with some special-effect makeup and primarily frontal faces. These datasets provide a comprehensive benchmark to assess model performance under varied real-world conditions. We compare our method with six state-of-the-art approaches: BeautyGAN\cite{beautygan}, PSGAN\cite{psgan}, EleGANt\cite{elegant}, CSD-MT\cite{csd-mt}, StableMakeup\cite{stableMakeup} and SHMT ($h_0$)\cite{sun2024shmt}. The first four are GAN-based, while StableMakeup and SHMT are diffusion-based architectures. CSD-MT and SHMT uses unsupervised learning and self-surpervised learning stragtegies while other methods use pseudo ground truth methods as mentioned in Related Work. Except for BeautyGAN, all methods utilize external face parsing models during pre-processing to enhance spatial feature representation.

\begin{figure*}[t!]
\begin{center}
\includegraphics[width=\textwidth]{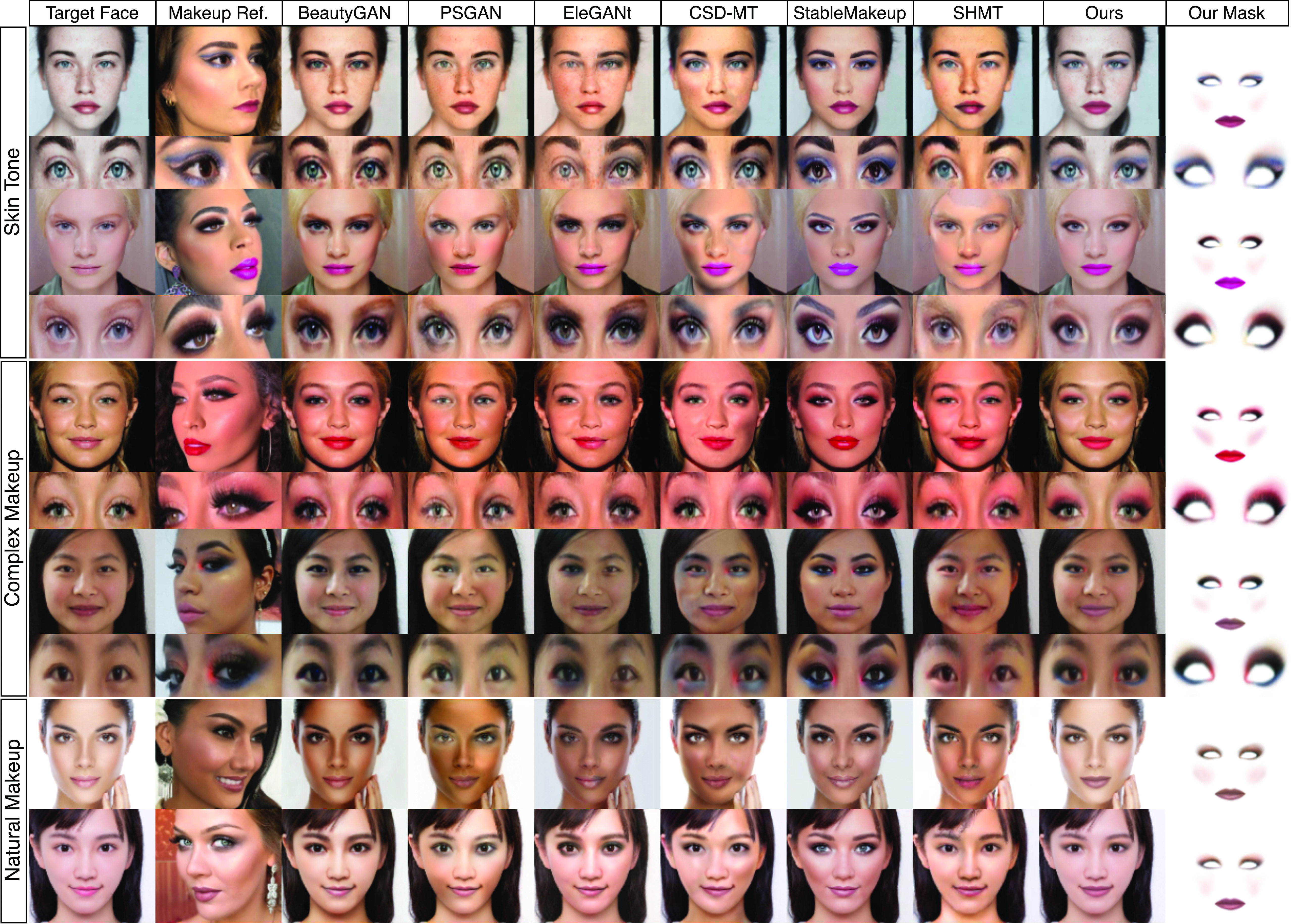}
\caption{\textbf{Qualitative comparison.} The makeup references are from Makeup-Wild dataset\cite{psgan} and the natural faces are from makeup transfer dataset\cite{beautygan}. Zoom in to see the makeup details.}
\label{fig:sota}
\end{center}

\end{figure*}
\begin{figure*}[]
\begin{center}
\includegraphics[width=\textwidth]{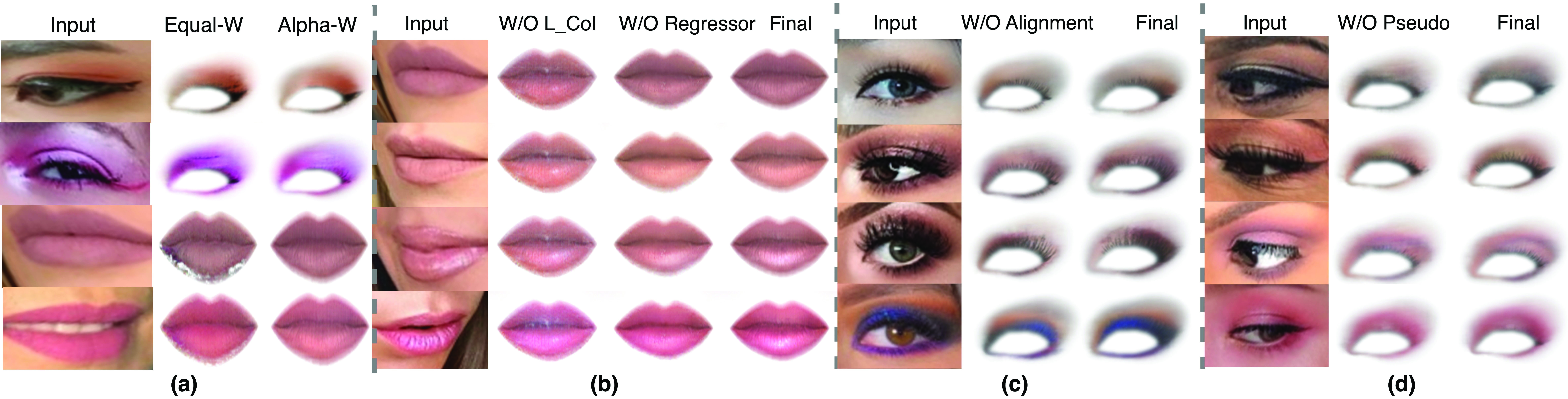}
\caption{Ablation study on Makeup-Wild dataset\cite{psgan}. (a) Comparison between equal-weighted and alpha-weighted reconstruction loss. (b) Comparison between direct color loss and color regressor supervision. (c) Effect of aligning input based on eye locations. (d) Impact of using k-means-based pseudo labels.}
\label{fig:eye_n_lip_ablation}
\end{center}
\end{figure*}

\subsection{Qualitative Comparison}
Figure \ref{fig:sota} compares our method with state-of-the-art approaches using the Makeup-Wild and MT datasets. Existing GAN-based methods often introduce shadow-like artifacts in cheek regions. GAN-based methods and SHMT struggle to preserve intricate, multi-color eye makeup. StableMakeup captures finer details and smoother cheeks but alters facial identity features, such as iris color, nose and lip structure, raising potential fairness concerns.
Our method effectively transfers fine-grained makeup while preserving identity integrity, achieving higher fidelity in complex makeup than existing GAN-based methods and more consistent identity preservation than StableMakeup. It also demonstrates robust transfer across diverse skin tones. Additional qualitative results are in the supplementary material.

\textbf{Video testing.} By decoupling makeup extraction and application, our method allows for extracting the makeup mask once and consistently applying it across entire video sequences. We test our pipeline using makeup images from the Makeup-Wild dataset and videos from the CelebV-Text dataset\cite{yu2022celebvtext}. As shown in Figure \ref{fig:front_page_vids}, our approach ensures consistent makeup application throughout the video. More video and video frame results are provided in the supplementary material.

\subsection{Quantitative Comparison}
We evaluate makeup transfer quality using three quantitative metrics: Fréchet Inception Distance (FID) \cite{heusel2017gans} for distribution-level perceptual similarity, Learned Perceptual Image Patch Similarity (LPIPS) \cite{lpips} for image-level perceptual similarity, and Peak Signal-to-Noise Ratio (PSNR) for pixel-level fidelity. Due to the lack of real before-and-after makeup image pairs, we adopt a synthetic evaluation strategy. We randomly sample two no-makeup faces at a time and apply makeup to both using the graphic-based approach described in section \ref{subsection:grphicdata}. Makeup is then transferred from the after-makeup image of the first face to the before-makeup image of the second face. The transferred result is compared to the after-makeup image of the second face. This process is repeated across all faces in the dataset. The evaluation results, reported in Table~\ref{tbl:FID}, show that our method outperforms all baselines across all metrics, demonstrating superior transfer fidelity.

To evaluate identity preservation, we perform transfer from real makeup reference images to no-makeup faces and compute the FID between the original faces and the transfer results. A lower FID in this context indicates stronger identity preservation under makeup modification. As shown in Table~\ref{tbl:FID}, our method achieves the lowest identity-preservation FID scores across all evaluated datasets, demonstrating its ability to decouple makeup from identity-relevant features.

\subsection{Ablation Study\label{subsection:ablation}}
\textbf{Alpha-weighted reconstruction loss. }To assess the alpha-weighted reconstruction loss, we train a baseline model with a equally-weighted reconstruction loss. As shown in Figure \ref{fig:eye_n_lip_ablation}, it fails to converge effectively, and produces artifacts in low-alpha regions, such as black line-like noise in eyeshadow and patchy lip edges.

\textbf{Lip color loss.} To assess the lip color loss on color accuracy, we train a baseline model without the lip color loss. As shown in Figure \ref{fig:eye_n_lip_ablation}, it often produces inconsistent lip coloration, including white artifacts between the lips in open-mouth cases and yellowish tones near the lip boundaries due to illumination variations in the reference images. In contrast, the model with the lip color loss achieves more uniform and accurate lip coloration, aligning more closely with the reference makeup. 

\textbf{Lip color regressor.} To evaluate whether the model can learn accurate lip color without relying on an explicit color regressor, we test an alternative training objective that directly minimizes the difference in average lip color between the prediction and the ground truth. Specifically, we replace Equation \ref{eq:L_col} with the new average color loss that does not use any color regressor, defined as:
{\small
\begin{equation}
L_{Col}^{NoReg}= \Big\vert\Big\vert \frac{1}{n}\sum^W_{i=1}\sum^H_{j=1}(M \odot \hat{y})-\frac{1}{n}\sum^W_{i=1}\sum^H_{j=1}(M \odot y)\Big\vert\Big\vert_2
\end{equation}
}
As demonstrated in Figure \ref{fig:eye_n_lip_ablation}, the direct color supervision results in overly smoothed lip textures. Notably, when comparing the predicted appearances of matte lipsticks in the first two rows and glossy lipsticks in the third and fourth rows, the makeup converges towards a uniform texture, exhibiting reduced variation in reflectance and finish. Moreover, the upper lip region appears blurred, indicating a loss of high-frequency details. These results suggest that the color regressor plays a critical role in improving overall color accuracy while preserving texture details.

\textbf{Eye alignment.} Figure \ref{fig:eye_n_lip_ablation} shows the effect of facial landmark-based alignment during reference image preprocessing. Without alignment, the model captures the dominant color clusters of the eyeshadow but frequently misplaces them spatially, for instance, confusing inner and outer eyeshadow regions. 

\textbf{Eye K-mean pseudo ground truth data.} As shown in Figure \ref{fig:eye_n_lip_ablation}, incorporating these K-mean pseudo ground truth data leads to improved rendering of fine-grained eye details. Upon closer inspection, we observe more precise outer eyeliner shapes and more consistent mascara density indicating that the pseudo ground truth help guide the model toward more accurate and detailed predictions.

\section{Limitations}
Our method relies on 2D facial keypoints and parsing for makeup application, which may lead to minor inaccuracies under extreme head poses or self-occlusion. While it performs well in typical scenarios, future work could explore dense 3D face alignment methods \cite{mediapipe_3dfacemesh} or generative image-to-image models to improve mask warping. Another limitation is the assumption of non-opaque, natural makeup styles. Our approach requires some visibility of the underlying skin tone to estimate transparency. Examples of out-of-distribution cases are provided in the supplementary material.

\section{Conclusion}
In conclusion, we propose a novel data generation pipeline and training strategy for extracting high-fidelity transparent makeup masks, along with an inference pipeline for temporally consistent, real-time makeup transfer. We introduce two complementary pseudo ground-truth generation methods: a graphics-based rendering approach and a k-means clustering technique. Our specialized training objectives further enhance the model’s ability to predict regionally accurate transparency masks. At inference time, we decouple makeup extraction from application, enabling consistent makeup transfer across video frames and supporting real-time rendering. Experimental results show that our method achieves high-fidelity, seamless virtual makeup try-on in videos and extends the applicability of makeup transfer across diverse skin tones.

\bibliography{aaai2026}

\appendix
\section{Technical Appendices and Supplementary Material}

\subsection{Network Structure Details}
For generate, we adapt the architecture from Pix2Pix~\cite{pix2pix} which is based on U-Net 256. We only change the number of input channels and output channels to four, as the model takes concatenation of RGB makeup reference and the alpha map as input, and outputs the RGBA makeup mask. The architecture of discriminator and color regressor are outlined in \ref{tbl:module_archi}. 

\begin{table}[ht]
\begin{center}
\begin{tabular}{@{}p{0.45\linewidth} p{0.05\linewidth} p{0.4\linewidth}@{}}
\toprule
\textbf{Color Regressor} & & \textbf{Discriminator} \\
\midrule
Conv2D(4,64,3,1,1)         & & Conv2D(7,64,3,1,1) \\
LeakyReLU(0.2)            & & LeakyReLU(0.2) \\
Conv2D(64,128,3,1,1)      & & Conv2D(64,128,3,1,1) \\
BatchNorm2D(128)          & & BatchNorm2D(128) \\
LeakyReLU(0.2)            & & LeakyReLU(0.2) \\
Conv2D(128,256,3,1,1)     & & Conv2D(128,256,3,1,1) \\
BatchNorm2D(256)          & & BatchNorm2D(256) \\
LeakyReLU(0.2)            & & LeakyReLU(0.2) \\
Conv2D(256,512,3,1,1)     & & Conv2D(256,512,3,1,1) \\
BatchNorm2D(512)          & & BatchNorm2D(512) \\
LeakyReLU(0.2)            & & LeakyReLU(0.2) \\
AdaptiveAvgPool2d((1,1))  & & Conv2D(512,1,3,1,0) \\
Linear(512,3)             & & Linear(36,18) \\
                          & & LeakyReLU(0.2) \\
                          & & Linear(18,1) \\
\bottomrule
\end{tabular}
\caption{Architecture for color regressor and discriminator. Conv2D parameters are: (input channels, output channels, kernel size, stride, padding); Linear layer parameters are: (input channels, output channels); AdaptiveAvgPool2d parameters are: ((image height, image weight)). Bias terms are disabled.}
\label{tbl:module_archi}
\end{center}
\end{table}

\subsection{Broader Impacts}
Previous makeup transfer methods primarily focus on preserving facial structure as the core aspect of identity, often overlooking skin tone as an important identity attribute. Consequently, when makeup is transferred between individuals with differing skin tones, the subject’s skin tone may be inadvertently altered. This not only affects identity preservation but also limits the applicability of makeup transfer technology across diverse demographic groups and raising fairness concerns. Our work addresses this limitation by explicitly disentangling makeup attributes from skin tone, ensuring that identity is preserved more holistically. By doing so, we aim to make makeup transfer more inclusive and accessible to a broader range of users, regardless of their skin tone.

Our method enables users to visualize realistic makeup effects before applying them in real life. However, excessive reliance on makeup transfer technology may reinforce unrealistic beauty standards and increase appearance-related anxiety.

\subsection{Illustration of Limitations}
\begin{figure*}[t!]
\begin{center}
\includegraphics[width=\textwidth]{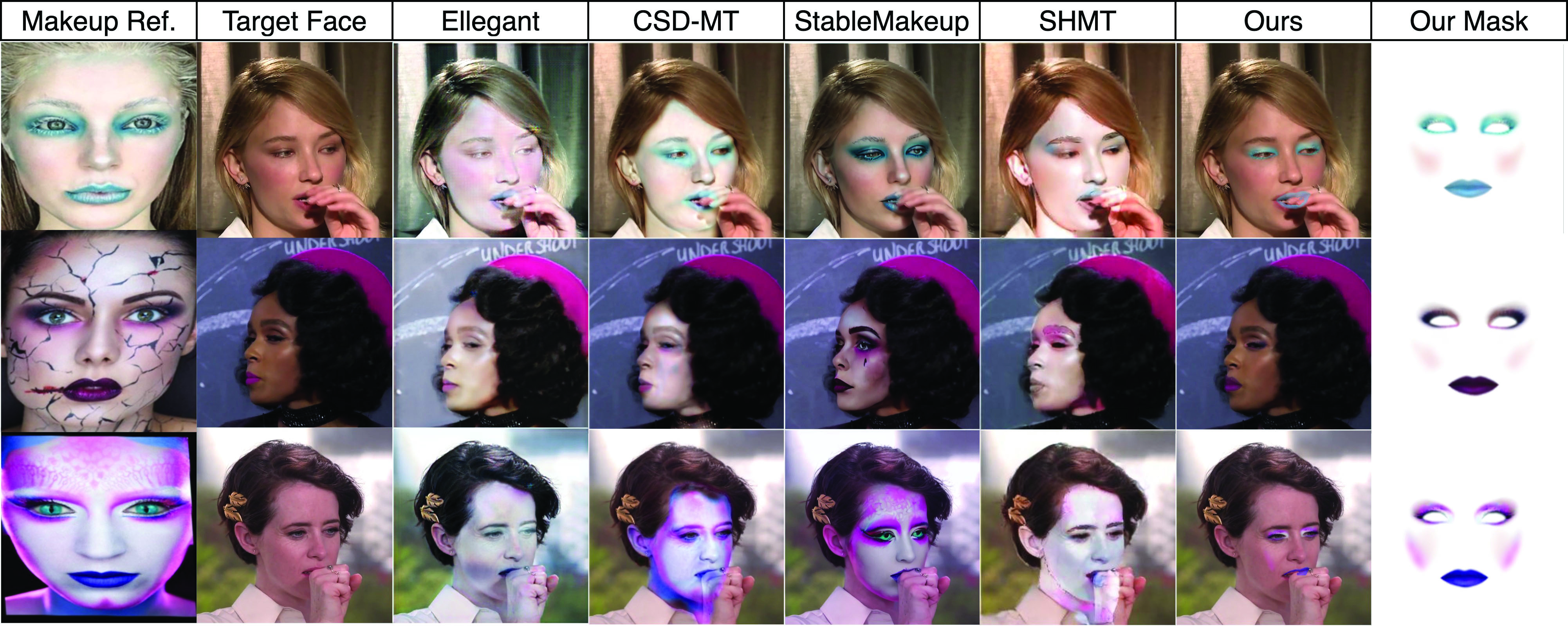}
\caption{Illustration of limitations. Our method relies on some visibility of the underlying skin tone to estimate makeup transparency and depends on 2D facial keypoints and parsing for accurate application. As a result, it is sensitive to landmark or parsing inaccuracies. Makeup references are from the LADN dataset~\cite{ladn}, and face images are from the CelebV-Text dataset~\cite{yu2022celebvtext}.}
\label{fig:failure_cases}
\end{center}
\end{figure*}

As shown in Figure~\ref{fig:failure_cases}, our method currently focuses on non-opaque and non-special effect makeup styles. While it relies on 2D facial landmarks and parsing, which may introduce minor inaccuracies under extreme poses or self-occlusion, it remains effective for most common use cases. Extending support to more complex makeup types and improving robustness to challenging head poses using 3D face alignment merhods are promising directions for future work.

\subsection{Additional Experiment Result}
\begin{figure*}[t!]
\begin{center}
\includegraphics[width=0.9\textwidth]{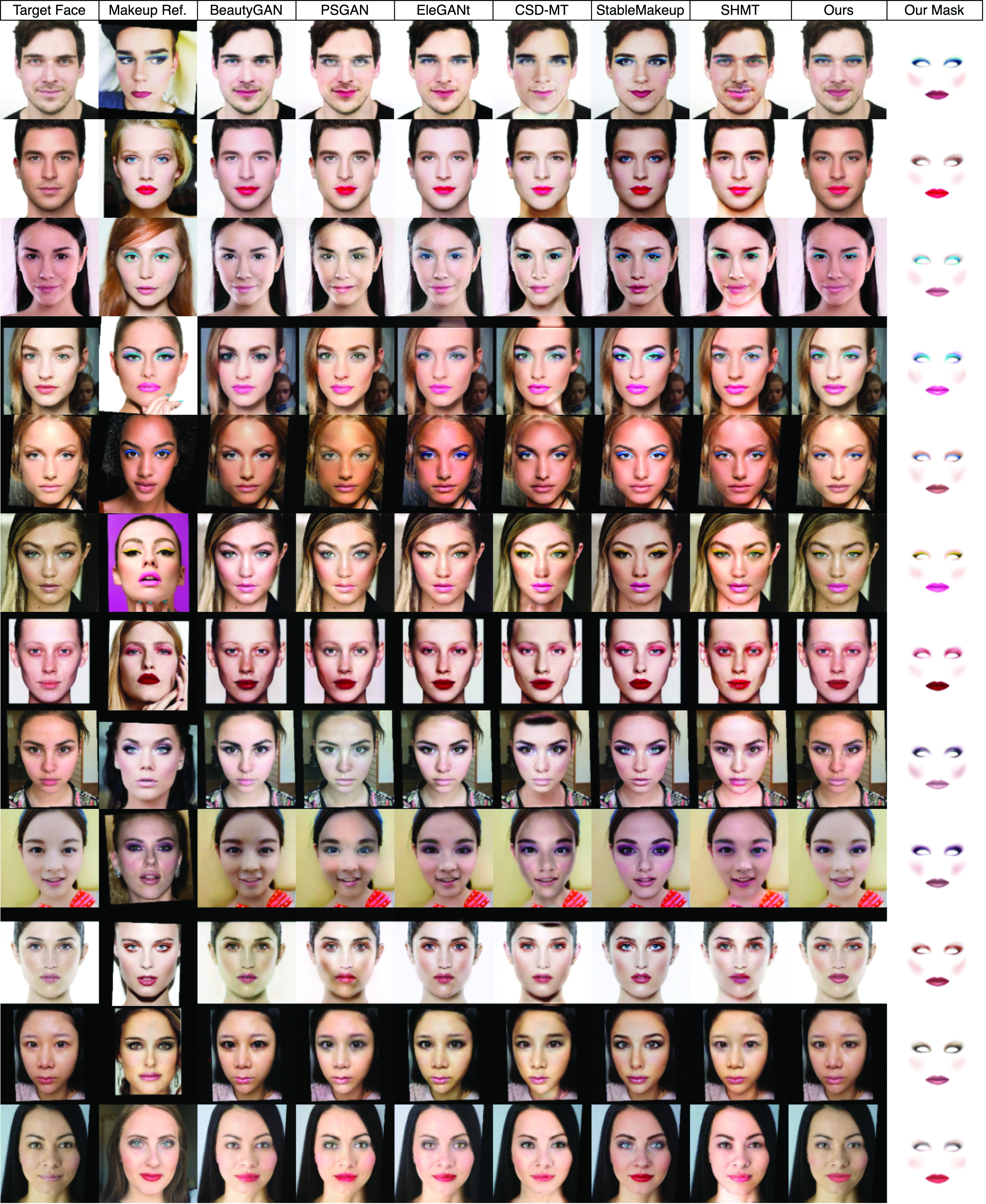}
\caption{Additional qualitative comparison with state of the art methods. The makeup references and faces are both from LADN dataset~\cite{ladn}.}
\label{fig:more_comparison_ladn}
\end{center}
\end{figure*}

More comprehensive qualitative comparison results on image data can be found in Figure \ref{fig:more_comparison} and Figure \ref{fig:more_comparison_ladn}.

\begin{figure*}[t!]
\begin{center}
\includegraphics[width=0.78\textwidth]{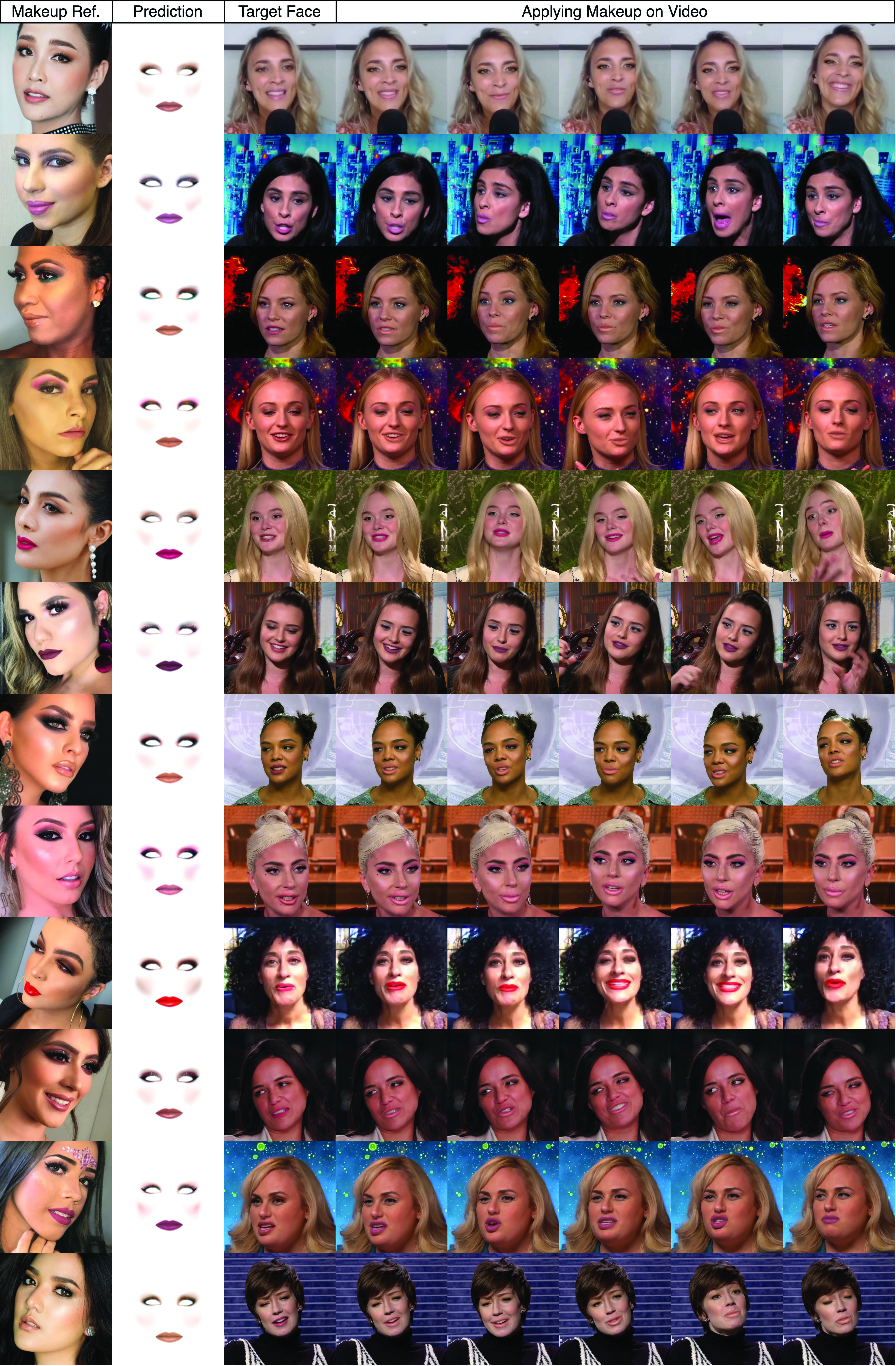}
\caption{Additional video testing result. The makeup reference images are from Makeup-Wild dataset~\cite{psgan} and videos are from CelebV-Text dataset~\cite{yu2022celebvtext}. More videos are in the Supplementary Material.}
\label{fig:supp_vids}
\end{center}
\end{figure*}

Additional video frame testing result can be found in Figure \ref{fig:supp_vids}. In each row, we extract the makeup mask from a given reference image and apply the same makeup mask throughout the video.

Additional video results are attached in the supplementary materials.

\subsection{Dataset Links}
The following datasets are used in our experiments can be found in the following links:

MT dataset~\cite{beautygan}: \url{https://github.com/wtjiang98/BeautyGAN_pytorch}

Wild-MT dataset~\cite{psgan}: \url{https://github.com/wtjiang98/PSGAN}

LADN dataset~\cite{ladn}: \url{https://github.com/wangguanzhi/LADN}

FFHQ dataset~\cite{styleGAN_ffhq}: \url{https://github.com/NVlabs/ffhq-dataset}

CelebV-Text dataset~\cite{yu2022celebvtext}: \url{https://github.com/celebv-text/CelebV-Text}

\subsection{Metrics Code Links}
The following implementation of metrics are used in our quantitative analysis.

We utilize the PyTorch official implementation of FID\cite{Seitzer2020FID}: \url{https://github.com/mseitzer/pytorch-fid}

We utilize the LPIPS calculation implemented by \cite{lpips}: \url{https://github.com/richzhang/PerceptualSimilarity}

See the metrics code in the supplementary material.

\subsection{Data Preprocessing Code Link}
We utilize the same preprocessing code implemented by \cite{elegant}: \url{https://github.com/Chenyu-Yang-2000/EleGANt/blob/main/assets/docs/prepare.md}

\begin{figure*}[t!]
\begin{center}
\includegraphics[width=\textwidth]{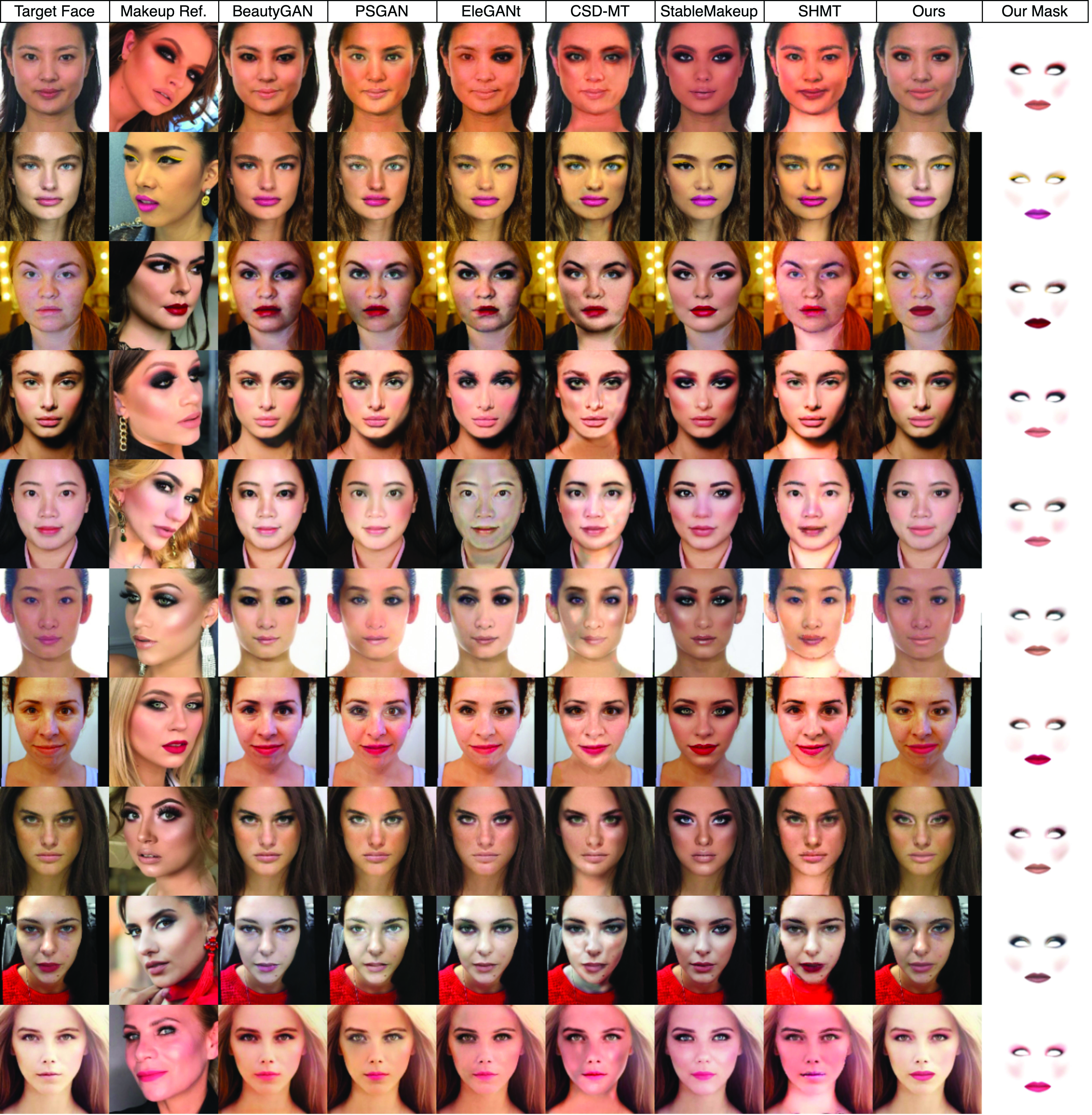}
\caption{Additional qualitative comparison with state of the art methods. The makeup references are from Makeup-Wild dataset~\cite{psgan} and the natural faces are from makeup transfer dataset~\cite{beautygan}.}
\label{fig:more_comparison}
\end{center}
\end{figure*}

\end{document}